# Social Learning Methods in Board Games

Vukosi N. Marivate and Tshilidzi Marwala

*Abstract*—This paper discusses the effects of social learning in training of game playing agents. The training of agents in a social context instead of a self-play environment is investigated. Agents that use the reinforcement learning algorithms are trained in social settings. This mimics the way in which players of board games such as scrabble and chess mentor each other in their clubs. A Round Robin tournament and a modified Swiss tournament setting are used for the training. The agents trained using social settings are compared to self play agents and results indicate that more robust agents emerge from the social training setting. Higher state space games can benefit from such settings as diverse set of agents will have multiple strategies that increase the chances of obtaining more experienced players at the end of training. The Social Learning trained agents exhibit better playing experience than self play agents. The modified Swiss playing style spawns a larger number of better playing agents as the population size increases.

## I. INTRODUCTION

INTELLIGENT agents [1] that play board games have been a focus in Machine Learning and Artificial Intelligence (AI). These agents are taught how to play games and learn from either saved games or by playing against themselves [2]. The problem that arises with agents that learn from playing against themselves is that they have a probability of not being able to capture all the dynamics of a game or variations in opponent's strategy. Thus self-play agents have a tendency to perform poorly against opponents that they have not come across their strategy before. To fix this, researchers have introduced the ability to save a large database of previously played games [2, 3]. This means that agents have databases of saved games that they can access. This results in large memory considerations as games increase their state sizes. This also increases computational complexity as searches within the databases are needed to find the best moves. Reinforcement learning [4] has been used extensively in multiple domains. One such domain has been in developing game playing agents. A problem that arises with self-play and reinforcement learning is the inability to model large state board games such as Go [2]. To deal with this problem this paper investigates the training of agents in social settings as opposed to self play and monitors the effects this has on the overall performance of the different agents created. This paper focuses on improving the performance of agents using a social setting. This is different from Social search/optimization methods such as Particle Swarm Optimization [5] or Memetic Algorithms [6]. The game playing agents are competitive, they are only trying to maximize their own performance and have no global goal. Meaning they are have no explicit knowledge of how well the whole social group is performing.

The paper first presents the background in Section II. Then the methodology is covered in Section III. Modeling of the game and testing is covered in Section IV and V respectively. Section VI presents the results and then the paper is concluded in Section VII.

## II. BACKGROUND

### A. Artificial Intelligence in Games

Making computers that have the ability to play games [2] against human opponents has been a challenge since the beginning of research into Artificial Intelligence in machines. Through the years there have been machines that have been taught to learn and play a multitude of games. Games that are currently mastered by machines, such that they cannot be beaten by humans, this includes; chess, backgammon, Othello and checkers [2] . Challenges, and their solutions, that arise in modelling of games can be extended to the real world. Problems associated with games are easier to model since they have rules that are bound and have event constraints. This is in contrast to real world problems where rules can change and there is a high level of uncertainty. The fact that games are simpler to model, does not imply that they cannot be useful in solving real world problems. The skills learnt by researchers in the field of AI in games, are helping them find new or improved solutions in multitudes of problems in other realms. In Reinforcement Learning [4][7] most agents learn to play games by playing games against themselves, termed self-play [2][8], for a large amount of iterations. Thus the agents learn from the experiences they create.

### B. Reinforcement Learning

An intelligent agent is defined [1] as a computer system/program that resides in some environment and is

V. N. Marivate is with the Computational Intelligence Research Group of the School of Electrical and Information Engineering, Private Bag 3, University of the Witwatersrand, Johannesburg, Wits, 2050. (phone: +27723292126; fax: +27865149542; e-mail: vukosi.marivate@ieee.org).

T. Marwala is with the Computational Intelligence Research Group of the School of Electrical and Information Engineering, Private Bag 3, University of the Witwatersrand, Johannesburg, Wits, 2050. (e-mail: t.marwala@ee.wits.ac.za).

allowed to perform actions in that environment. Humans learn by interacting with each other. Lessons are learned from being rewarded or punished after performing an action. This is different from supervised learning [3]. In supervised learning, a learning algorithm is given test cases that have inputs and the corresponding correct outputs. This for example, can be in the form of function approximation as shown in equation (1).

$$\bar{y} = f(\bar{x}) \qquad (1)$$

Where *x* can be a vector of multiple inputs and *y* is a vector that is composed of multiple outputs. Thus the learning algorithm tries to approximate the function f(.). Reinforcement learning can be categorized as unsupervised learning. An agent is placed in an environment. It performs actions in that environment and perceives the effects of the actions in that environment through its sensors/receptors. The agent also receives a reward/punishment given the change the action has made in the environment. This reward can be extrinsic (from the environment) or intrinsic (from within the agent) [9]. This is illustrated in Figure 1.

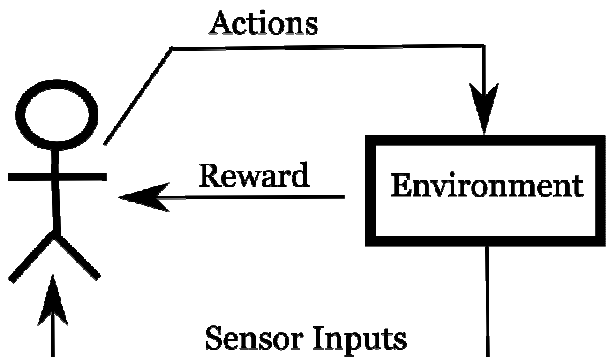

Figure 1. Reinforcement Learning Framework

In a general reinforcement learning problem one deals with a Markov Decision Problem [4] (MDP). An MDP is made up of a number of entities.

- *S* - set of states of the environment
- *s* - current state of the environment
- *s'* - The next state
- *A* - set of actions that can be taken by the agent
- *a* - current action chosen by the agent
- R – Reward given *(R(s), (R(s,a), R(s,a,s'))*
- $P(s'|s,a)$ -Transitional Probability

The transitional probability is the probability of moving into another state (*s'*) given an action (*a*) and a state (*s*). Given the above information, an agent can make a decision on which actions are best to take in a specific state. This is termed the policy (π) of the agent. It is a mapping of a state to a specific action (*a=π(s)*). The transitional probabilities of an environment are not normally provided or known. Thus a challenge in reinforcement learning is modelling an environments dynamics within the agent. To do this the concept of the value of a state is introduced. This is done through the introduction of Value Function and Action Value functions. Through these functions one can evaluate the policy that the agent is taking. The value function is defined in (2) as:

$$V_\pi(s) = E_\pi[\sum_{k=0}^{\infty} \gamma^k r_{t+k+1} \mid s_0 = s] \qquad (2)$$

This is the expected value (*E*) of the summation of the discounted (γ) reward (*r*) of all possible future states given that the agent is executing a policy π given that we are starting at the current state. The policy (π) is the mappings of state to actions. The action-value function is

$$Q_\pi(s,a) = E_\pi[\sum_{k=0}^{\infty} \gamma^k r_{t+k+1} \mid s_0 = s, a_0 = a] \qquad (3)$$

Where *Q(s,a)* takes into account not only beginning at the the current state but also the current action. The maximization of (2) and (3) by carrying out an optimal policy π* will result in higher rewards in the end. To find the policy that maximises the value function or action-value function we use the Bellman Optimality equations [4]. To learn in reinforcement learning from a system without the complete model (Model free) of the system then the agent needs to learn through experience. The agent thus has to go through interactions and find an optimal policy that optimizes (2) or (3).

*C. Learning Algorithm and Action Selection*

The learning algorithm used in this paper is the TD-Lambda Algorithm [4]. The algorithm is applied to action value functions as in (3). The algorithm allows the agents initially to explore and as they play more games start exploiting more and exploring less. In this paper, action value functions are used with a table structure. Function approximation is not used and the toolbox used for modeling and implementing the experiments is the Reinforcement Learning toolbox built by Gerhard Neumann [10].

The generalised form of a terminal difference algorithm combines bootstrapping like dynamic programming methods [4] and sampling like Monte-Carlo methods. The algorithm is shown in Figure 2.

```
Initialise V(s) to an arbitrary value and e(s) = 0
Repeat (for each game)
    Initialise s
    Repeat (for each game step)
        a ← action given for π for s
        Take action a, observe reward, r, and next state, s'
        δ ← r + γV(s')-V(s)
        e(s) ← e(s)+1
        For all s:
            V(s) ← V(s)+αδe(s)
            e(s) ← γλe(s)
        s ← s'
    until end of game
```

Figure 2. TD-Lambda Algorithm

$e(s)$ above is the eligibility trace [11] of a certain state. Thus if a certain state repeats itself its update is taken into account with a higher importance depending on how recent the previous occurrence was.

For choosing the actions and allowing exploration and exploitation actions were chosen using and epsilon greedy distribution which can be written as:

$$P(s,a_i) = \begin{cases} 1-\varepsilon + \frac{\varepsilon}{|A|}, & \text{if } a_i = \arg\max_{a' \in A} Q(s,a') \\ \frac{\varepsilon}{|A|}, & \text{else} \end{cases} \quad (4)$$

The agent chooses a random action with probability ε and takes a greedy action with probability 1 – ε. This makes sure that the agents initially are more likely to explore but as more and more games are played ε decreases and thus the agents start to then exploit more by using the knowledge that they have gained through playing the games.

*D. Social Learning*

As reinforcement learning develops from modeling how humans develop in their early stages of life another complimentary theory can be used in conjunction with reinforcement learning. Humans seldom learn only by themselves. They live in a society and thus observe what others do. This is termed social learning. In gaming circles this is even more distinct. Players of such board games such as chess, Scrabble and checkers mentor each other in their clubs [14]. In social learning there are a number of important factors that a being must have in order to be able to learn. The being or in this case agent must be able to [12]:

- Pay attention to the what is being observed
- Remember the observations
- Be able to replicate the behavior
- Be motivated to demonstrate what they have learnt

Thus learning by observing involves four processes: attention, retention, production and motivation. In reinforcement learning this can be extended to being able to play a game and observe state transitions, remember what actions have been taken, trying a different action after previous one failed and then being rewarded if it leads to a terminal state. Further an agent then observes what an opponent does. Vygotsky [13] discusses the concept of the more knowledgeable other. This concept takes into account that in a social setting an agent would learn more from another agent who has more experience or is at the same level. This can also be observed in chess clubs where members are paired to train with stronger players or peers.

By introducing other agents as opponents in the learning stage one introduces a non-stationary playing environment [14]. If for example the opponent is a logic based intelligence computer program, a reinforcement learning agent would learn a strategy or policy that would optimally beat the logic opponent [15]. Thus stimulating a social setting is needed. This would then increase the probability of creating agents that not only just know how to beat a specific opponents strategy but has a broader knowledge of a state space. This is discussed further in the proceeding sections.

III. METHODOLOGY

Humans play and learn board games in groups. This community of players imparts knowledge on each other. If one looks at communities of chess or Scrabble [16] players one can see that very experienced players mentor weaker players. To simulate a social learning environment such as this, multiple agents need be created. In this paper each agent is given its own identity in that they have different initialization parameters. The agents have the same learning algorithm but have different initialization options. This is shown in Table 1.

TABLE I
AGENT IDENTITIES

| Parameter | Range |
|---|---|
| Learning Rate | 0.2 – 0.3 |
| Discount Factor | 0.95 -0.99 |
| Lamda | 0.9-1.0 |

Two training configurations are used in training the agents in the social setting. The two methods are derived from tournament styles. A modified Swiss [17] and a Round Robin system are used and compared. In the modified Swiss configuration, agents are paired up to play one round of a game which is a full episode. When the game is finished there is either a winner or a loser or there is a draw. A tournament like structure was utilised for the agents to play in. The structure is shown in Figure 3.

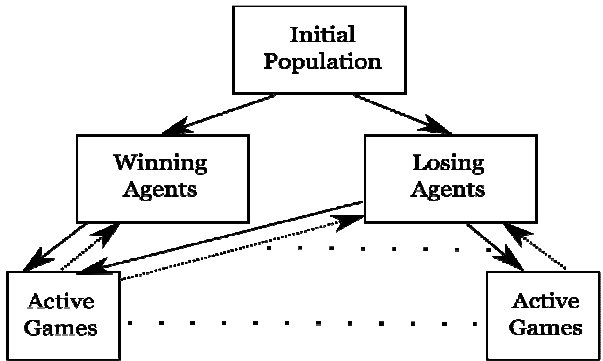

Figure 3 Tournament Learning Framework

The agents are first initialized and placed in an initial population. In the first iteration they are arbitrarily put in two sub-classes (Winning Agents and Losing Agents). In the second iteration and for the rest of the game the agents play games against each other. A winning agent is pitted against a losing agent. After a game/episode the winner is placed in the winner agent list and the losing agent in the losing agent list, thus a direct simulation of a mentor and a learner. At the end of a playing round the agents will be in two groups. A number of rounds are played and the process of pairing losers and winners repeats until the maximum number of rounds is reached. In this configuration there is a large focus on getting agents to be paired with players that have better experience.

In a round robin setting each agent plays against the other. There is no splitting of the group to winners and losers. After a round of playing the players are then pitted against the next player. This is done until the maximum number of games is played. This has less of a focus on having a more knowledgeable other or a peer as an opponent. Another agent was created which is the self-play agent. This agent learns by only playing against itself. It plays a move as one player and then plays another move as the other player. This agent was created so as to be able to benchmark how well the social agents fair against conventional self-play learning.

## IV. MODELLING THE GAME AND LEARNING

### A. Tic Tac Toe

Tic-Tac-Toe [18] is a 3 x 3 board game. Two players place pieces on the board trying to connect three of their own pieces in a row. Figure 4 illustrates the player with the noughts defeating the player with the crosses.

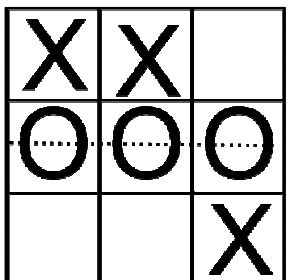

Figure 4. Tic Tac Toe Board

If two great players play a game of Tic-Tac-Toe it should always end with a draw [2]. The game has been modeled with reinforcement learning in the past [5]. It has been recorded that agents take 50000 learning episodes [19] to be able to play at a beginner level. In this experiment this is the amount of iterations used for the training of the agents.

### B. The Game Model

To model the game for reinforcement learning the game was represented by 10 state variables. Nine of the variables can have 3 different values which represent the places on the board. Each place on the board can be empty or have a nought or cross. The tenth state is the current player who is supposed to play. The model also keeps track of which actions are available to an agent in a certain state. Thus an illegal move such as placing a piece on a board area that already has a piece is not possible. When an agent wins a game it is rewarded with a reward of 1.0. When the agent loses it then gets a reward of -1.0. When there is a draw, the agent gets a reward of 0.0. For all other game states that are not terminal the reward is 0.0.

### C. Learning

The games are managed by a game controller. The controller allocates who has to play next and also keeps track of game statistics such as wins, test results and how many times each agent has played games. It also matches winners and losers and thus implements the social frameworks described in section III. The agents are initialized with different learning parameters. Thus the agents play against non-stationary opponents. This stimulates the emergence of more robust agents. The opponents policies are also changing and thus a learner will have to adjust its policy to be a policy that can play against more than one stationary opponent.

## V. TESTING

### A. Board Test

Two tests were setup for the agents. The first test for the agents was an assessment on how well the agents perform at trying to pick correct actions in given test states. The Tic-Tac-Toe board is setup with pieces already on it. There is only one correct move that can be made. There were a total of 10 test boards with different levels of difficulty. The agents are given one try at each board. Some boards have to reach a terminal state (end of game) while in others the agent has to choose an action that will result in forcing a draw in the game. There are 5 easy boards, 2 intermediate boards and 3 hard boards. The easy boards test if the agent can notice states that will make them win (Figure 5).

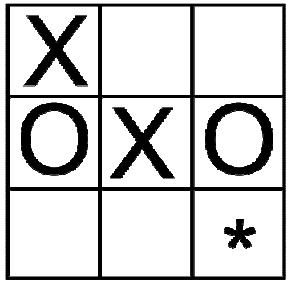

Figure 5. Cross to Play (Easy)

These are 1 move to win boards. They are relatively easy and test how the agents try to choose actions that will maximum reward in their next action choice. The intermediate boards are defensive boards where they test how well an agent can block a win by the other opponent, which means a loss for the agent, or force a draw. These tests show that the agent is trying to avoid losing or getting a lower return. An example is illustrated in Figure 6.

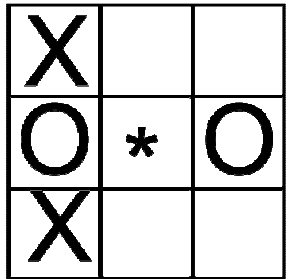

Figure 6. Cross to Play (Intermediate)

The difficult boards test how an agent can force a win his future move and not the next move. These are trickier but test how the agent is trying to maximize its future returns. The board is shown in Figure 7.

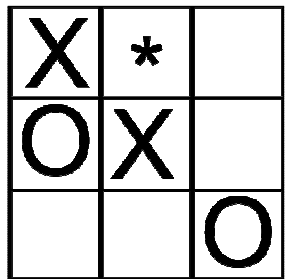

Figure 7. Cross to move (Hard)

### B. Play Test

The second test the agents take is taking part in a league. All of the agents are allowed to play with all the other agents. The wins, losses and draws are recorded. This is used to find which of the agents are the strongest. 5000 games are played by the agents against each other. This was applied to the best modified Swiss agents and Self-Play agents.

### C. Testing Method

The agents were built with different population sizes. The first size is 4, then 6 and then 8. Each of these was tested 5 different times with the board test (meaning they have been trained differently 5 times) and then 5 times with the play test. The results are presented in the following section.

## VI. RESULTS

### A. Board Test Results

The tests were carried through with different agent populations. The results of the tests for the modified Swiss configuration are shown below in Figure 8.

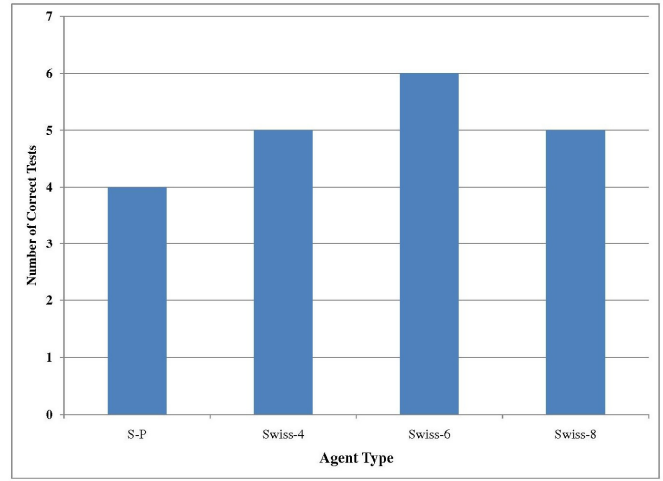

Figure 8. Board test results SP vs. Swiss Self Play

The results show that the Self-Play(S-P) agent gets 4 moves correct while the best Swiss social agent in the 4 population (SO4) gets 5 while the one in the 6 (SO6) gets 6 correct. This implies that the Self play agent plays at a beginner level while the SO6 is playing at an intermediate level compared to the other agents. None of the agents are advanced.

The other test was with the Round Robin Configuration. The results are in shown in Figure 9.

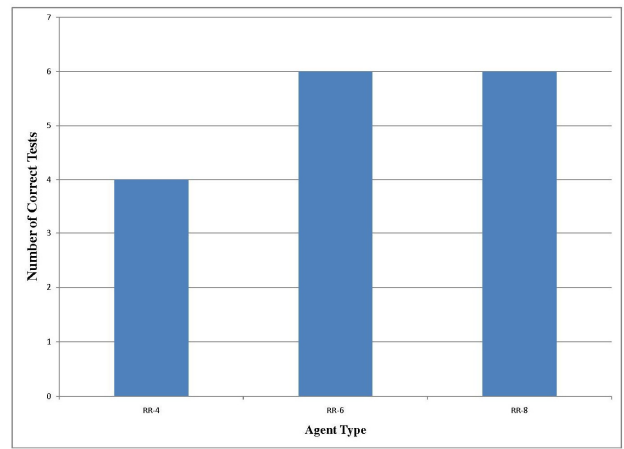

Figure 9. Round Robin Agent Performance

Another observation from the social agents is that as more and more agents (>8) are used in the population there is an increase in the number of intermediate agents in one generation. This is more evident in the Swiss tournament setting as opposed to the Round Robin configuration. Both configurations were tested with 16 and 32 agent sized populations. When the populations are increased with the modified Swiss configuration more than one intermediate agent emerges. In some stages up to 6 intermediate agents emerge. With the Round Robin configuration 2 intermediate playing agents have emerged.

By introducing multiple different agents as opponents in the training phases, one has been able to create agents that are superior to the S-P agent.

### B. Play Test Results for Modified Swiss

The play test was conducted on the Swiss Configuration social agents. A sample of the board test results is shown in Table 2.

TABLE 2
AVERAGE SWISS SOCIAL AGENT PLAY TEST

|     | S-P  | SO1  | SO2  | SO3  | SO4  |
| --- | ---- | ---- | ---- | ---- | ---- |
| S-P | 0    | 3049 | 3041 | 3028 | 3046 |
| SO1 | 3084 | 0    | 3058 | 3077 | 3089 |
| SO2 | 3024 | 3034 | 0    | 3063 | 3080 |
| SO3 | 3060 | 3059 | 2995 | 0    | 3047 |
| SO4 | 3037 | 3063 | 3028 | 3022 | 0    |

All of the agents played 5000 games against each other. In the above table there are 4 social agents and one self-play agent. The self play agent is the best agent that was kept during initialization and training of the agents. Thus the best agent that performed in the board tests is used. In the above table the S-P agent won 3041 games against SO, while SO1 won 3084 games against. Thus the difference is 43 games more that were won by SO1. This indicates that when an agent starts a game first they are more likely to win. The agents on average in the above configuration are winning over 60 % of the games they start first. This shows the agents still have weaknesses in defending. This is expected as the agents are all playing at a very low level. The social agents on average beat the self-play agents 50 times or more.

## VII. CONCLUSION

The agents all play the game at beginner level. This is indicated by how they perform at the board test. All of the agents fare very well on the easy boards but struggle on the intermediate ones and the difficult ones. There are a number of intermediate agents that are created in the social settings. Thus without increasing the number of training cycles, but by introducing non-stationary opponents in social settings the agent's performance have been improved. The larger the population sizes the more likely the number of superior agents. In this paper a small number was used with positive results and it is expected that with large population sizes the agents will have better performance increases. This would be a mimic of real world populations of players where you have thousands of players in any sport.

In the play tests the beginner level of the agents is further shown as they all have higher chances of winning if they start the game first. The social agents have made it possible to create agents that are superior to the best self-play agents. This is a positive result and merits the potential for the use of social methods in agent learning.

ACKNOWLEDGMENT

V. N. Marivate thanks Simon Jagoe for his assistance with setting up the environments as well as Gerhard Neumann with the assistance with the setup of the toolboxes.